\documentclass{article}
\usepackage{ijcai19}

\usepackage{times}
\usepackage{soul}
\usepackage{url}
\usepackage[hidelinks]{hyperref}
\usepackage[utf8]{inputenc}
\usepackage[small]{caption}
\usepackage{graphicx}
\usepackage{amsmath}
\usepackage{booktabs}
\usepackage{multirow}
\usepackage{algorithm}
\usepackage{algorithmic}
\usepackage{bm}
\usepackage{ dsfont }
\usepackage{amssymb}
\DeclareMathOperator*{\argmax}{argmax}
\usepackage{amsthm}
\usepackage{color}
\newtheorem{definition}{Definition}
\newtheorem{theorem}{Theorem}
\newtheorem{corollary}{Corollary}

\title{Cascaded Algorithm-Selection and Hyper-Parameter Optimization with Extreme-Region Upper Confidence Bound Bandit~\thanks{This work is supported by the National Key R\&D Program of China (2017YFB1001903), NSFC (61876077), Jiangsu SF (BK20170013), and Collaborative Innovation Center of Novel Software Technology and Industrialization. Yang Yu is the corresponding author.}}

\author{
Yi-Qi Hu
\and
Yang Yu
\and
Jun-Da Liao
\affiliations
National Key Laboratory for Novel Software Technology, Nanjing University, Nanjing 210023, China
\emails
\{huyq, yuy\}@lamda.nju.edu.cn,
liaojd98@gmail.com
}

\begin{document}

\maketitle

\begin{abstract}
	An automatic machine learning (AutoML) task is to select the best algorithm and its hyper-parameters simultaneously. Previously, the hyper-parameters of all algorithms are joint as a single search space, which is not only huge but also redundant, because many dimensions of hyper-parameters are irrelevant with the selected algorithms. In this paper, we propose a cascaded approach for algorithm selection and hyper-parameter optimization. While a search procedure is employed at the level of hyper-parameter optimization, a bandit strategy runs at the level of algorithm selection to allocate the budget based on the search feedbacks. Since the bandit is required to select the algorithm with the maximum performance, instead of the average performance, we thus propose the extreme-region upper confidence bound (ER-UCB) strategy, which focuses on the extreme region of the underlying feedback distribution. We show theoretically that the ER-UCB has a regret upper bound $O\left(K \ln n\right)$ with independent feedbacks, which is as efficient as the classical UCB bandit. We also conduct experiments on a synthetic problem as well as a set of AutoML tasks. The results verify the effectiveness of the proposed method.
	
\end{abstract}

\section{Introduction}

    Algorithm selection and hyper-parameter optimization are core parts of automatic machine learning (AutoML). Previously, AutoML approaches often define the search space as the algorithm selection space~\cite{modelselect3,modelselect1,modelselect2}, hyper-parameter space~\cite{hu2018expracos,hu2019multi}, or the joint of the both spaces (CASH problem)~\cite{autosklearn-nips,autoweka-kdd}. While the joint space allows a more thorough search that could cover potentially better configurations, the huge space is a barrier to effective search in limited time. Moreover, the joint space can be quite redundant when considering only one of the algorithms, since the hyper-parameters of the other algorithms are irrelevant. Therefore, the joint space contains redundancy or even can be misleading. 
    
    The cascaded algorithm selection can have levels~\cite{jamieson2016non}. The first level is on the hyper-parameter optimization. It only needs to focus on the selected algorithm, but not the hyper-parameters of all algorithms. The second level is on the algorithm selection. However, previous methods in this kind commonly carry out a full hyper-parameter optimization on the candidate algorithms, making the slow and expensive algorithm evaluations. 
    
    In this paper, we propose a cascaded algorithm selection approach to avoid a full-space hyper-parameter optimization. The hyper-parameter optimization usually employs some stepping search methods, which can be paused after every search step, and can also be resumed. The selection receives feedback and allocates the next search step to one of the algorithms. Thus, the cascaded algorithm selection is naturally to be modeled as a multi-armed bandit problem~\cite{auer2002finite}. However, most of the classical bandits maximize the average feedbacks. In the AutoML, however, only the best feedback matters. A variant of the bandit, the \emph{extreme bandit}~\cite{carpentier2014extreme}, can model this situation, which tries to identify the arm with the maximize (or equivalently minimize) feedback value. However, as the extreme bandit follows the extreme distribution, it is not only unstable but often require to known the distribution type, making the extreme bandit approach unpractical.
    
    In this paper, we propose the \emph{extreme-region UCB bandit} (ER-UCB), which focuses on the extreme region of the feedback distributions. Unlike the extreme bandit, ER-UCB considers a region instead of the extreme point, which can lead to a better mathematical condition. Moreover, in machine learning where the test data is commonly different from the train data, the extreme region can be more robust for generalization. With $K$-arms and $n$ trials, our analysis proves that ER-UCB has the $O\left(K \ln n\right)$ regret upper bound, which has the same order with the classical UCB strategy. The experiments on synthetic and real AutoML tasks reveal that the ER-UCB can find the best algorithm precisely, and exploit it with the majority of the trial budget.
    
    The rest sections present background \& related works, extreme-region UCB bandit, experiments, and conclusion.

\section{Background \& Related Works}

    We consider the algorithm selection and hyper-parameter optimization on classification tasks. Let $\mathcal{D}^{\text{train}}$ and $\mathcal{D}^{\text{test}}$ denote the training and testing datasets. Let $\bm{C}=\{C_1, C_2,\dots, C_K\}$ denote the algorithm set with $K$ candidates. For $C_i \in \bm{C}$, $\bm{\delta}_i \in \Delta_i$ denotes a hyper-parameter setting, where $\Delta_i$ is the hyper-parameter space of $C_i$. Let $f\left(\cdot\right)$ denote a performance criterion for a configuration $C_i^{\bm{\delta}_i}$, e.g., accuracy, AUC score, etc. The AutoML problem can be formulated as follows:
    \begin{align}
        C_{i^*}^{\bm{\delta_i^*}}=\argmax_{C_i \in \bm{C}, \bm{\delta}_i \in \Delta_i} \frac{1}{k} \sum_{j=1}^{k} f\left(C_i^{\bm{\delta}_i}, \mathcal{D}_{j}^{\text{train}}, \mathcal{D}_{j}^{\text{valid}}\right),
    \end{align}
    where $\mathcal{D}_{j}^{\text{valid}} \in \mathcal{D}^{\text{train}}$ and $\mathcal{D}_{j}^{\text{train}}=\mathcal{D}^{\text{train}}-\mathcal{D}_{j}^{\text{valid}}$. It is also concludes the CASH problem formulation~\cite{autosklearn-nips}.
    
    Because of the non-convex, non-continuous and non-differentiable properties, derivative-free optimization~\cite{sac-aaai,sracos-aaai} is usually applied to solve it. For example, a tree-structure based Bayesian optimization (SMAC)~\cite{smac-lion} is employed on AutoWEKA~\cite{autoweka-kdd} and AutoSKLEARN~\cite{autosklearn-nips}, the popular open-source AutoML tools. Derivative-free optimization explores search space by sampling and evaluating. But the high time-cost restrains the total number of evaluations on AutoML. With the limited trials, the performance of derivative-free optimization is very sensitive to search space. However, in above formulation, the search space $\Delta=\Delta_1 \times \Delta_2, \times, \dots, \times \Delta_K$. Obviously, $\Delta$ is redundant, because the best configuration is only relevant to the hyper-parameter space of the best algorithm.
    
    Hence, we consider an easier formulation, i.e., optimizing hyper-parameters of algorithms separately:
    \begin{align}
        \label{automl-target}
        & C_{i^*}^{\bm{\delta^*}} = \argmax_{C_i \in \bm{C}} \frac{1}{k} \sum_{j=1}^{k} f\left(C_i^{\bm{\delta}_{i}^{*}}, \mathcal{D}_{j}^{\text{train}}, \mathcal{D}_{j}^{\text{valid}}\right), \nonumber \\ 
        & \text{where, } \bm{\delta}_{i}^{*} = \argmax_{\bm{\delta}_i \in \Delta_i} \frac{1}{k} \sum_{j=1}^{k} f\left(C_i^{\bm{\delta}_{i}}, \mathcal{D}_{j}^{\text{train}}, \mathcal{D}_{j}^{\text{valid}}\right).
    \end{align}
    The hyper-parameter processes can be seen as arms. The algorithm selection level is a multi-armed bandit problem. The bandit is a classical formulation of the resource allocation problem. In~\cite{felicio2017multi}, the authors formulated the cold-start user recommendation as a multi-armed bandit problem, which user information was unavailable at the beginning. The feedbacks of users has to be obtained by trials. In this situation, the bandit concerns more about the average feedback of arms. In~\cite{cicirello2005max}, the authors proposed the max $K$-armed bandit, which focused on the maximum feedback of trials. But it assumed that the reward distribution was a Gaussian distribution, and it was designed for the heuristic search, in which more than one arms can be selected at a trial step. 
    
    In this paper, we customize the  extreme-region UCB (ER-UCB) bandit for AutoML problems.

\section{Extreme-Region UCB Bandit}

    In this section, we present details of the ER-UCB: the bandit formulation for AutoML, the deduction of the ER-UCB strategy and the theoretical analysis on the ER-UCB strategy.
    
    \subsection{Bandit formulation for AutoML}
        
    In the classical multi-armed bandit, feedbacks of an arm obey an underlying distribution. In this paper, we employ the random search on the hyper-parameter optimization. A trial in a model $C_i$ is uniformly sampling hyper-parameters from $\Delta_i$, and its performance is the feedback of this trial. Thus, $X_i \sim \mathfrak{D}_i\left(\mu_{X_i}, \sigma_{X_i}^2\right)$, where $X_i$ denote a feedback of a trial on $C_i$, and $\mathfrak{D}_i\left(\mu_{X_i}, \sigma_{X_i}^2\right)$ is the underlying performance distribution of $C_i$. Because of the random search, $\mathfrak{D}_i$ is fixed. With $K$ algorithm candidates, let $\bm{\mathfrak{D}}=\{\mathfrak{D}_1, \mathfrak{D}_2, \dots, \mathfrak{D}_K\}$ denote the performance distribution set. The $K$-armed bandit formulation for AutoML is: at the $t$-th trial, the $C_{I_t}$ is selected from $K$ algorithm candidates, and get a feedback $X_{I_t}$ independently from $\mathfrak{D}_{I_t}$. 

    \subsection{Deduction}
    
    In AutoML tasks, the selected algorithm is required to have maximum performances. For this requirement, we present the extreme-region target for the proposed bandit. Then, we show the deduction details of extreme-region UCB strategy.
    
    \subsubsection{Extreme-region target}
    The target of the hyper-parameter optimization is to find the hyper-parameters which have the  maximum performance. In the bandit, with a fixed $\epsilon$, we want the probability $\text{Pr}\left[X_i \geq \mu_{X_i} + \epsilon\right]$ as large as possible. With the Chebyshev inequality: $\text{Pr}\left[X_i \geq \mu_{X_i} + \epsilon\right] \leq \frac{\sigma_{X_i}^{2}}{\epsilon^2}$, let $\frac{\sigma_{X_i}^{2}}{\epsilon^2}=\theta$,
    \begin{align}
        \text{Pr}\left[X_i \geq \mu_{X_i} + \sqrt{\frac{1}{\theta}}\sigma_{X_i}\right] \leq \theta.
    \end{align}
    In other words, with the same fixed probability upper bound $\theta$, the best arm selection is:
    \begin{align}
        \label{truth_selection}
        I_{t}=\argmax_{i \in \{1,2,\dots,K\}} \mu_{X_i} + \sqrt{\frac{1}{\theta}}\sigma_{X_i}.
    \end{align}
    With the given $\mu_{X_i}$ and $\sigma_{X_i}$, the ground-truth selection strategy is (\ref{truth_selection}). But, when facing the unknown distributions, we have to estimate the expectation and variance based on the observations. With the Markov inequality, it is easy to relate the expectation $\mu_{X_i}$ with its estimation. But for variance, it is hard to find the relationship. With the variance definition:
    \begin{align}
        \label{exp_equation}
        \mathbb{E}\left[X_{i}^{2}\right] & = \sigma_{X_i}^{2} + \mu_{X_i}^{2}. 
    \end{align}
    Because $\mathbb{E}\left[X_{i}^{2}\right]$ is the expectation of the random variable $X_i^2$. The Markov inequality can be applied to it easily. And $\mathbb{E}\left[X_{i}^{2}\right]$ can partly represent $\sigma_{X_i}^{2}$ according to (\ref{exp_equation}). Thus, we try to replace $\sigma_{X_i}$ with $\sqrt{\mathbb{E}\left[X^{2}_i\right]}$:
    \begin{align}
        \label{middle_selection}
        I_{t}=\argmax_{i \in \{1,2,\dots,K\}} \mu_{X_i} + \sqrt{\frac{1}{\theta}\mathbb{E}\left(X_i^2\right)}.
    \end{align}
    Comparing with (\ref{truth_selection}), (\ref{middle_selection}) magnifies the effect of expectation item on selection strategy. To tackle this issue, we introduce a hyper-parameter $\beta \geq 0$, and construct a new random variable $Y_i=X_i-\beta$. Furthermore, let $\mu_{Y_i}=\mathbb{E}\left[Y_i\right]$,  $Z_i=\left(X_i-\beta\right)^2$ and $\mu_{Z_i}=\mathbb{E}\left[Z_i\right]$. Thus, the extreme-region target is:
    \begin{align}
    \label{final_selection}
    I_{t}=\argmax_{i \in \{1,2,\dots,K\}} \mu_{Y_i} + \sqrt{\frac{1}{\theta}\mu_{Z_i}}.
    \end{align}
    We prove that it can reduce the effect of expectation on algorithm selection by introducing into $\beta$:
    \begin{proof}
        According to definitions of $X_i$, $Y_i$ and $Z_i$,
        \begin{align}
            \mu_{Y_i} + \sqrt{\frac{1}{\theta}\mu_{Z_i}} = \mu_{X_i} -\beta  + \sqrt{\frac{1}{\theta}\left(\sigma_{X_i}^{2}+\left(\mu_{X_i}-\beta\right)^2\right)}.
        \end{align}
        Comparing with (\ref{truth_selection}), because of $\beta \geq 0$, the item of expectation is reduced, but the item of variance stays the same. It concludes the proof.
    \end{proof}
    
    \subsubsection{Extreme-region UCB strategy}
    We apply the upper confidence bound (UCB) strategy on the extreme-region target. In this paper, we assume that the random variables satisfy the following moment condition. There exists a convex function $\psi$ on the reals, for all $\lambda \leq 0$,
    \begin{align}
        \label{assume1}
        \ln \mathbb{E} \  e^{\lambda|X-\mathbb{E}\left[X\right]|} \geq \psi\left(\lambda\right).
    \end{align}
    If we let $X \in \left[0, 1\right]$ and $\psi\left(\lambda\right)=\frac{\lambda^2}{8}$, (\ref{assume1}) is known as Hoeffding's lemma. We apply this assumption to construct an upper bound for the estimated expectations at some fixed confidence level. Let $\psi^*$ denote the Legendre-Fenchel transform of $\psi$. With $s$ observations of $X_i$, let $\hat{\mu}_{Y_i}^{s}=\frac{1}{s}\sum_{t=1}^{s}Y_{i,t}$ and $\hat{\mu}_{Z_i}^{s}=\frac{1}{s}\sum_{t=1}^{s}Z_{i,t}$ denote the estimated expectations of $Y_i$ and $Z_i$. Only for $Y_i$ with a fixed $\epsilon_{Y_i}$, using the Markov inequality:
    \begin{align}
        \label{equation_y}
        \text{Pr}\left[\mu_{Y_i} \geq \hat{\mu}_{Y_i}^{s} + \epsilon_{Y_i}\right] \leq e^{-s\psi^{*}\left(\epsilon_{Y_i}\right)}.
    \end{align}
    The same deduction for $Z_i$, and $f\left(x\right)=\sqrt{x}$ is a monotonically increasing function:
    \begin{align}
        \label{equation_z}
        \text{Pr} \left[\sqrt{\frac{1}{\theta}\mu_{Z_i}} \geq \sqrt{\frac{1}{\theta}\left(\mu_{Z_i}^{s}+\epsilon_{Z_i}\right)}\right] \leq e^{-s\psi^{*}\left(\epsilon_{Z_i}\right)}.
    \end{align}
    Because $\sqrt{a+b}\leq\sqrt{a}+\sqrt{b}$, and let $\epsilon_{Y_i}=\epsilon_{Z_i}=\epsilon$. With the union bound,  we combine $Y_i$ and $Z_i$ as follows:
    \begin{align}
        \text{Pr}\Bigg[\mu_{Y_i} + \sqrt{\frac{1}{\theta}\mu_{Z_i}} \geq \hat{\mu}_{Y_i}^{s}  + & \sqrt{\frac{1}{\theta}\hat{\mu}_{Z_i}^{s}} + \epsilon+\sqrt{\frac{1}{\theta}\epsilon}\Bigg]  \nonumber \\
         & \leq 2e^{-s\psi^{*}\left(\epsilon\right)}.
    \end{align}
    Let $2e^{-s\psi^{*}\left(\epsilon\right)}=\delta$. With the probability at least $1-\delta$,
    \begin{align}
        \hat{\mu}_{Y_i}^{s} + & \sqrt{\frac{1}{\theta}\hat{\mu}_{Z_i}^{s}} +  \left(\psi^*\right)^{-1}\left(\frac{1}{s}\ln\frac{2}{\delta}\right)  \nonumber \\ 
        & + \sqrt{\frac{1}{\theta}\left(\psi^*\right)^{-1}\left(\frac{1}{s}\ln\frac{2}{\delta}\right)} > \mu_{Y_i}+
        \sqrt{\frac{1}{\theta}\mu_{Z_i}}.
    \end{align}
    Within total $s$ trials, let $T_{i}\left(s\right)=\sum_{t=1}^{s}\mathds{1}_{I_t=i}$ denote the number that the $i$-th arm is selected, and $\frac{2}{\delta}=t^{\alpha}$. $\left(\alpha, \psi\right)$-ER-UCB strategy is:
    \begin{align}
        \label{alpha-ucb}
        I_t = & \argmax_{i \in \{1,2,\dots,K\}} \Omega_i\left(Y_i, Z_i, T_i\left(t\right)\right)+\Psi_i\left(T_i\left(t\right), t\right), \text{where, }\nonumber \\
         & \Omega_i=\hat{\mu}_{Y_i}^{T_i\left(t\right)}+ \sqrt{\frac{1}{\theta}\hat{\mu}_{Z_i}^{ T_i\left(t\right)}}, \nonumber \\
        & \Psi_i=\left(\psi^*\right)^{-1}\left(\frac{\alpha \ln t}{T_{i}\left(t\right)}\right) + \sqrt{\frac{1}{\theta}\left(\psi^*\right)^{-1}\left(\frac{\alpha \ln t}{T_{i}\left(t\right)}\right)}.
    \end{align}
    $\Omega_i$ and $\Psi_i$ are the exploitation and exploration items. With Hoeffding's lemma, taking $\psi\left(\lambda\right)=\frac{\lambda^2}{8}$, then, $\psi^*\left(\epsilon\right)=2\epsilon^2$. And let $\alpha=4$. The exploration can be re-written as: 
    \begin{align}
        \label{hoeffding-ucb}
        \Psi^{'}_i = \sqrt{\frac{2\ln t}{T_i\left(t\right)}} + \sqrt{\frac{1}{\theta}\sqrt{\frac{2\ln t}{T_i\left(t\right)}}}.
    \end{align}
    Thus, the Hoeffding's ER-UCB strategy is:
    \begin{align}
        I_t = & \argmax_{i \in \{1,2,\dots,K\}} \Omega_i\left(Y_i, Z_i,T_i\left(t\right)\right)+\Psi^{'}_i\left(T_i\left(t\right), t\right).
    \end{align}
    Because $X_i \in \left[0, 1\right]$ on AutoML, the exploitation item is often much smaller than the exploration item. To further exploration and exploitation trade-off, we introduce a hyper-parameter $\gamma$. The practical Hoeffding's ER-UCB strategy is:
    \begin{align}
        \label{practical_strategy}
        I_t = & \argmax_{i \in \{1,2,\dots,K\}} \gamma\Omega_i\left(Y_i, Z_i,T_i\left(t\right)\right)+\Psi^{'}_i\left(T_i\left(t\right), t\right).
    \end{align}
    \begin{algorithm}[tb]
        \caption{Extreme-region UCB Bandit}
        \label{er-ucb-alg}
        \textbf{Input}: ~~\\
        {\color{white} inp} $\{C_1, C_2,\dots, C_K\}$: $K$ model candidates;\\
        {\color{white} inp} $\{\Delta_1, \Delta_2, \dots, \Delta_K\}$: hyper-parameter spaces of models;\\
        {\color{white} inp} $\gamma,\ \theta,\  \beta$: hyper-parameters;\\
        {\color{white} inp} $n$: trial budget; \\
        {\color{white} inp} $\mathcal{D}^{\text{train}}$: train dataset of task;\\
        {\color{white} inp} $\texttt{Sample}_{\mathcal{U}}$: uniform sample sub-procedure;\\ 
        {\color{white} inp} $\texttt{Evaluate}$: evaluation sub-procedure. \\
        \textbf{Procedure}:
        \begin{algorithmic}[1] 
            \FOR {$t=1$ to $K$}
            \STATE $X_t=\texttt{Evaluate}\left(C_t, \texttt{Sample}_{\mathcal{U}}\left(\Delta_t\right), \mathcal{D}^{\text{train}}\right)$
            \STATE $\hat{\mu}_{Y_t}=X_t-\beta, \  \hat{\mu}_{Z_t}=\left(X_t-\beta\right)^2, T_t=1$
            \ENDFOR
            \FOR {$t=K+1$ to $n$}
            \STATE get index $I_t$ according to (\ref{practical_strategy})
            \STATE $X_{I_t}=\texttt{Evaluate}\left(C_{I_t}, \texttt{Sample}_{\mathcal{U}}\left(\Delta_{I_t}\right), \mathcal{D}^{\text{train}}\right)$
            \STATE $\hat{\mu}_{Y_{I_t}}=\frac{T_{I_t}\left(t\right)\hat{\mu}_{Y_{I_t}}+X_{I_t}-\beta}{T_{I_t}\left(t\right)+1}$
            \STATE $\hat{\mu}_{Z_{I_t}}=\frac{T_{I_t}\left(t\right)\hat{\mu}_{Z_{I_t}}+\left(X_{I_t}-\beta\right)^2}{T_{I_t}\left(t\right)+1}$
            \STATE $T_{I_t}=T_{I_t}+1$
            \ENDFOR
            \RETURN the hyper-parameters with the best $X$.
        \end{algorithmic}
    \end{algorithm}

    The cascaded algorithm selection and hyper-parameter optimization with ER-UCB bandit is presented at Algorithm~\ref{er-ucb-alg}. Line 2 and 7 are the procedures of uniformly sampling hyper-parameters for the selected algorithm and obtaining the feedbacks. Line 1 to 4 are the initialization steps. In the main loop (line 5 to 10), the algorithm is selected by the ER-UCB strategy (line 6). Line 7 to 9 are the procedures for updating the exploitation item for the selected algorithm.
    
    We have to discuss the hyper-parameters, i.e., $\theta$, $\gamma$ and $\beta$ for the ER-UCB bandit. $\theta$ is employed to control the space size of the extreme region. It is usually a small real number, e.g., 0.1 or 0.01. $\gamma$ is the exploration-and-exploitation trade-off hyper-parameter. In AutoML tasks, $\gamma$ is used to magnify the exploitation item. Thus, it is usually a big number such as 10 or 20. $\beta$ is applied to reduce the impact of expectation item in the selection strategy. It should be tuned according to tasks. In experiments, we will investigate them empirically. 
    
    \subsection{Theoretical Analysis}
    
    We present the analysis of the upper bound for $\left(\alpha, \psi\right)$-ER-UCB strategy (\ref{alpha-ucb}) and the Hoeffding's ER-UCB strategy (\ref{hoeffding-ucb}) on the extreme-region regret. For the arbitrary arm $i$ and a fixed $\rho$, we define $\text{Pr}\left[X_i \geq \rho \right]=p_i$. Thus, $p^{*} = \argmax_{i \in \{1,2,\dots,K\}} p_i$. According to (\ref{final_selection}), let $i^{*}=\argmax_{i \in \{1,2,\dots,K\}} \mu_{Y_i}+\sqrt{\frac{1}{\theta}\mu_{Z_i}}$, thus $\mu_{Y}^{*}+\sqrt{\frac{1}{\theta}\mu_{Z}^{*}}=\mu_{Y_{i^*}}+\sqrt{\frac{1}{\theta}\mu_{Z_{i^{*}}}}$, and $\Gamma_i=\mu_{Y}^{*}+\sqrt{\frac{1}{\theta}\mu_{Z}^{*}} - \mu_{Y_{i}} - \sqrt{\frac{1}{\theta}\mu_{Z_{i}}}$. We assume $p^{*}=p_{i^{*}}$ by choosing an appropriate $\beta$. The extreme-region regret is the Definition~\ref{regret-def}.
    \begin{definition}[Extreme-region regret]
        \label{regret-def}
        At $n$-th trial, event A is the number of times that $X_{i^*} \geq \rho$ occurs, and event B is the number of times that $X_{I_t} \geq \rho$ occurs with a given strategy. The extreme-region regret is:
        \begin{align}
            \text{R}_{n} = np^*-\mathbb{E}\sum_{t=1}^{n}p_{I_{t}}. \nonumber
        \end{align}
    \end{definition}
    Introducing $T_{i}\left(s\right)$ and $\Theta_{i}=p^{*}-p_i$, The extreme-region regret can be re-written as:
    \begin{align}
        \text{R}_n & = \left(\sum_{i=1}^{K}\mathbb{E}T_i\left(n\right)\right)p^*-\mathbb{E}\sum_{i=1}^{K}T_i\left(n\right)p_i \nonumber \\
        & = \sum_{i=1}^{K}\Theta_{i}\mathbb{E}T_i\left(n\right).
    \end{align}
    We can prove the following simple upper regret bound for $\left(\alpha,\psi\right)$-ER-UCB strategy:
    \begin{theorem}[Regret of $\left(\alpha,\psi\right)$-ER-UCB]
        \label{theorem1}
        Assume the feedback distribution of arbitrary arm satisfy (\ref{assume1}). With $\alpha > 2$, $\left(\alpha,\psi\right)$-ER-UCB satisfies:
        \begin{align}
            \text{R}_n \leq \sum_{i:\Gamma_{i}>0} \Theta_{i} \left(\frac{\alpha \ln n}{\psi^*\left(\Gamma_{i}^2/\left[4\left(1+\theta^{-1}\right)^2\right]\right)}+\frac{\alpha+2}{\alpha-2}\right). \nonumber
        \end{align}
    \end{theorem}
    Due to the limitation of paper length, we present the proof details in our supplementary material. Based on Theorem~\ref{theorem1}, we can easily prove the extreme-region regret of the Hoeffding's ER-UCB strategy:
    \begin{corollary}[Regret of Hoeffding's ER-UCB]
        Assume the feedback distribution of arbitrary arm satisfy (\ref{assume1}). With $\alpha > 2$, Hoeffding's ER-UCB satisfies:
        \begin{align}
            \text{R}_n \leq \sum_{i:\Gamma_{i}>0} \Theta_{i} \left(\frac{8\alpha \ln n}{\Gamma_i^{4}/\left(1-\theta^{-1}\right)^{4}}+\frac{\alpha+2}{\alpha-2}\right). \nonumber
        \end{align}
    \end{corollary}
    According to the theoretical analysis, the ER-UCB bandit has $O\left(K \ln n\right)$ upper bound on the extreme-region regret.
    
    \begin{figure*}[!t]        
        \centering
        \begin{minipage}{0.24\linewidth}\centering
            \includegraphics[width=\textwidth]{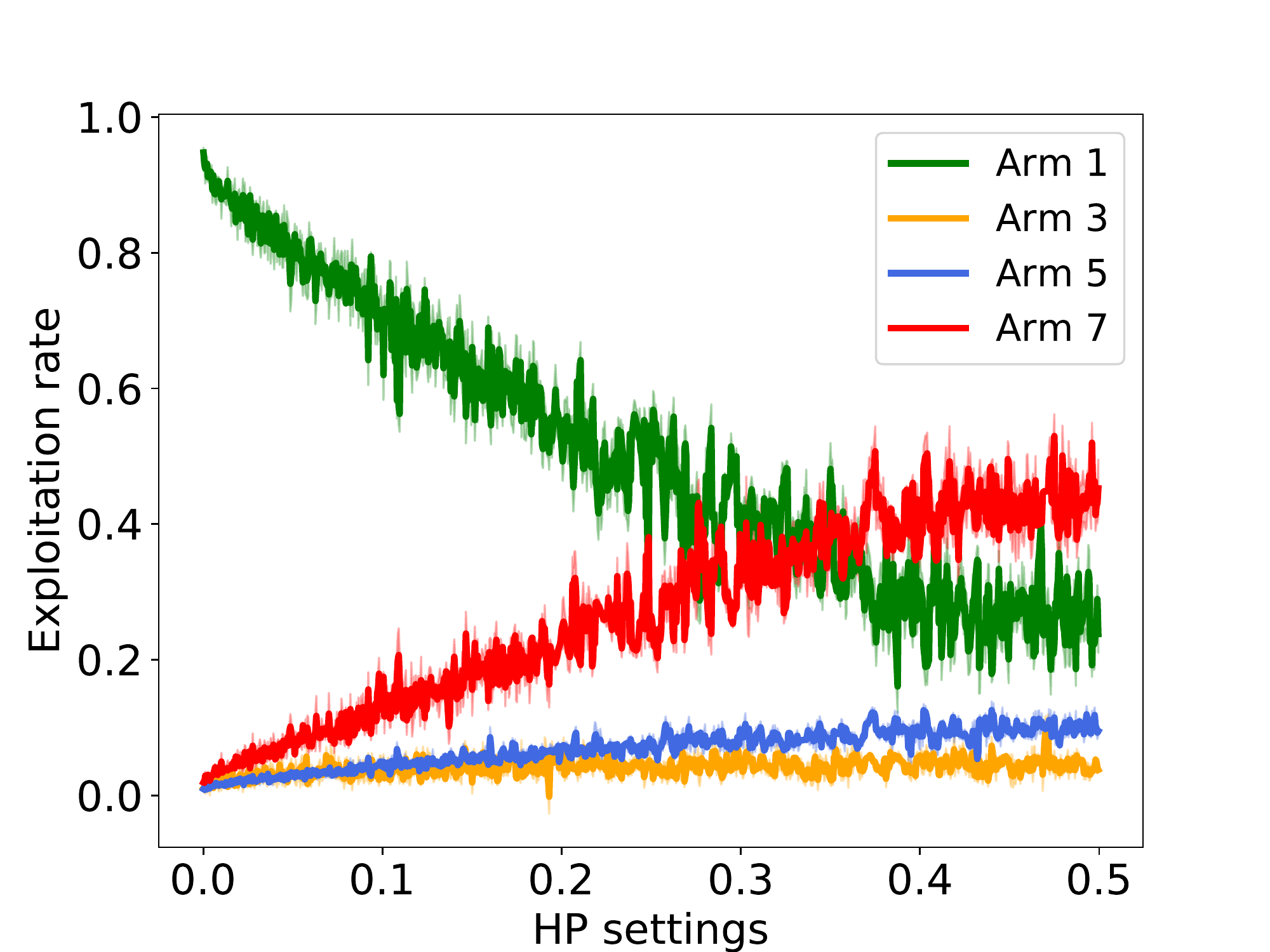}\\
            a.1 HP study for $\theta$ 
        \end{minipage} 
        \begin{minipage}{0.24\linewidth}\centering
            \includegraphics[width=\textwidth]{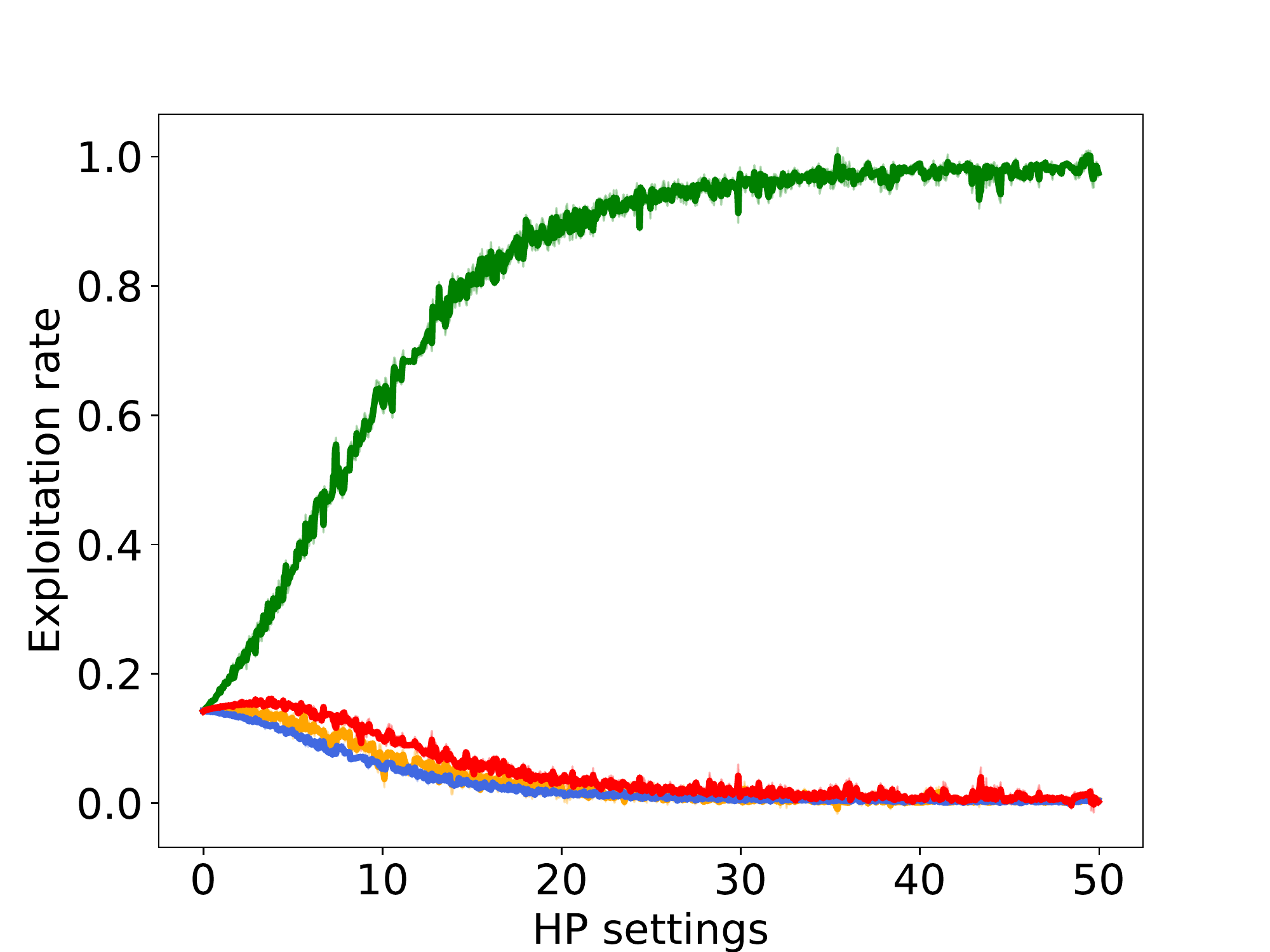}\\
            a.2 HP study for $\gamma$ 
        \end{minipage} 
        \begin{minipage}{0.24\linewidth}\centering
            \includegraphics[width=\textwidth]{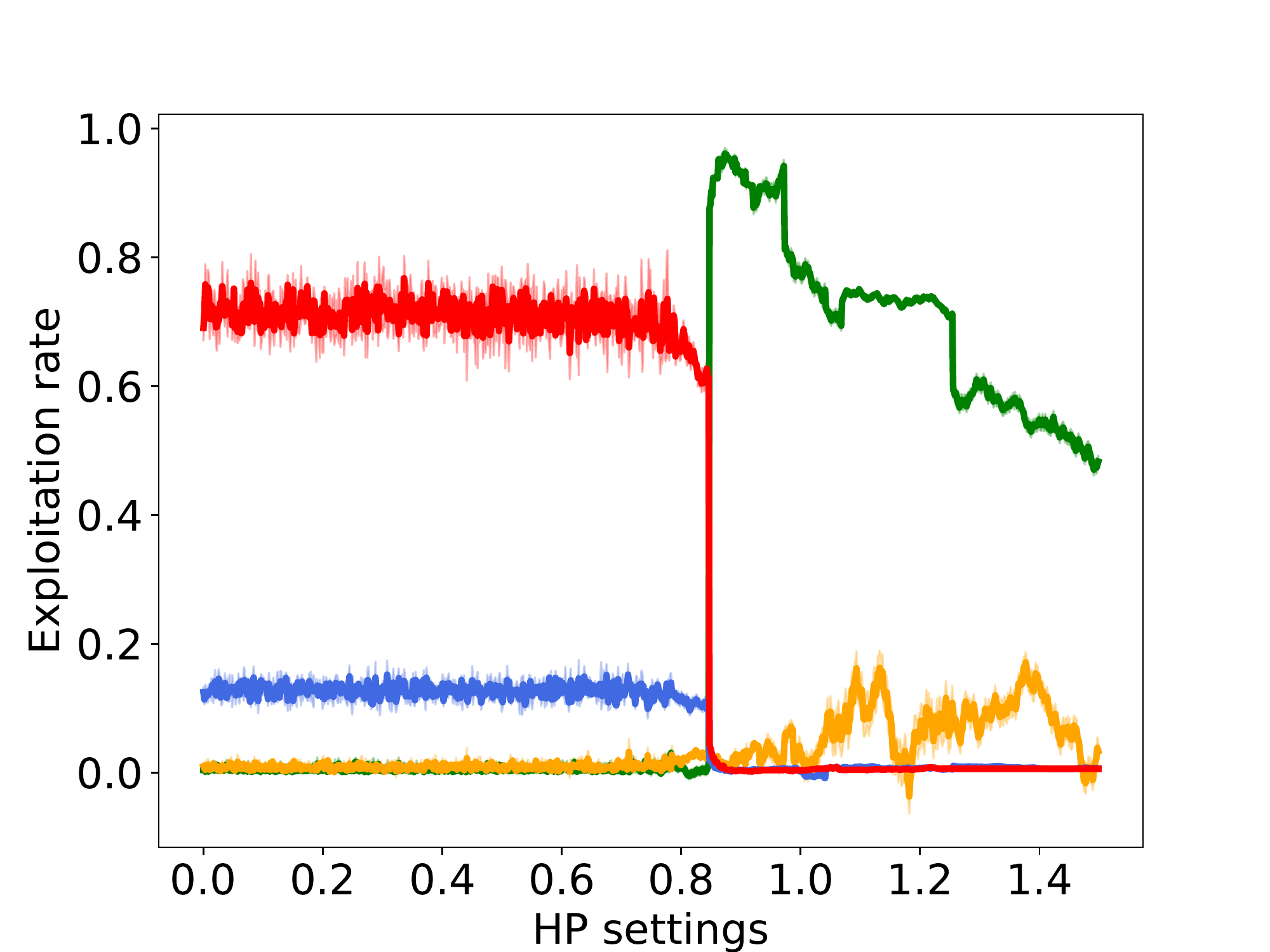}\\
            a.3 HP study for $\beta$
        \end{minipage}
        \begin{minipage}{0.24\linewidth}\centering
            \includegraphics[width=\textwidth]{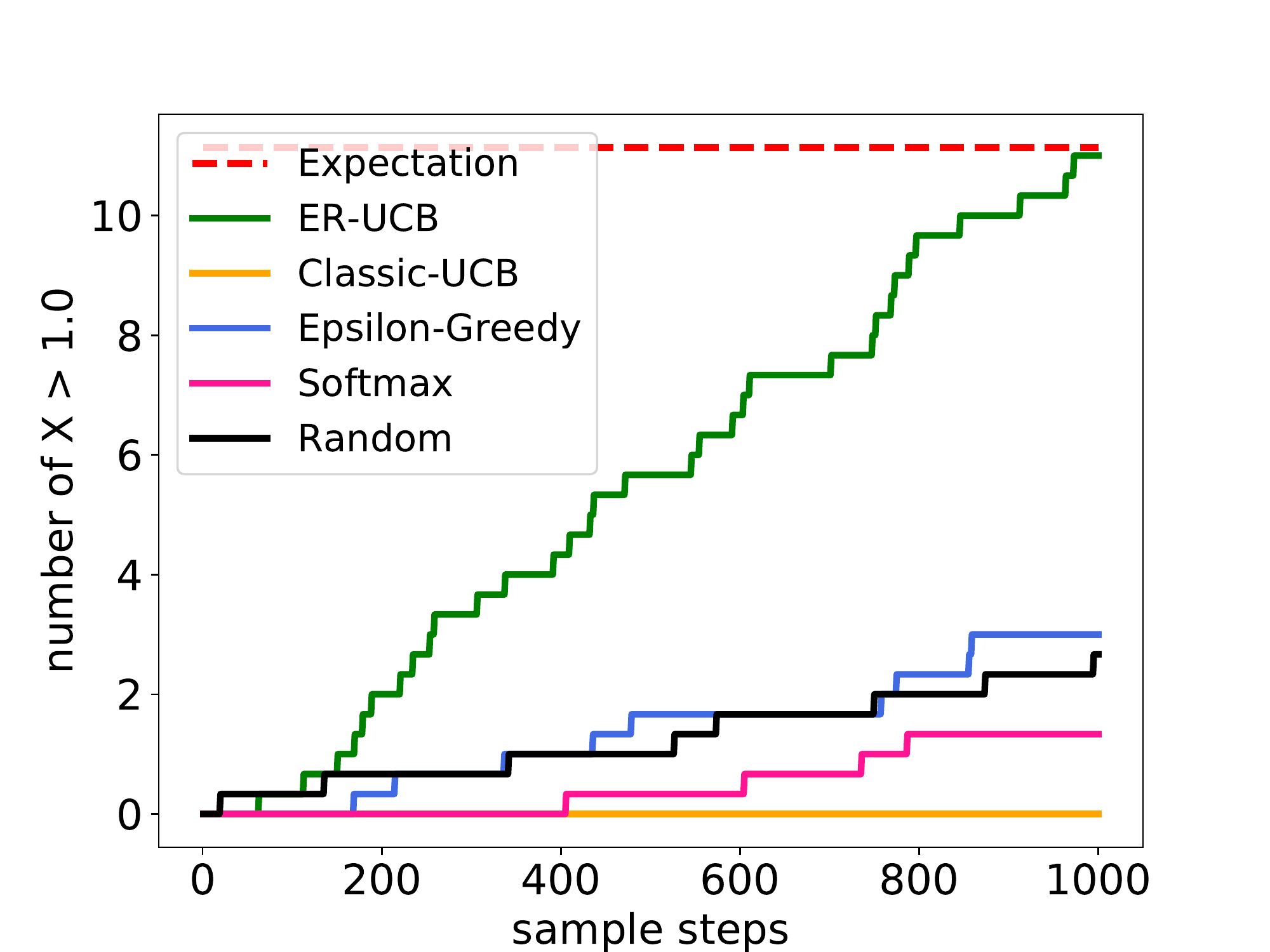}\\
            b. Regret study
        \end{minipage} 
        \caption{Illustrations of the results for the synthetic experiment. Figure a.1, a.2, a.3 are the results of the ER-UCB hyper-parameter studies, which illustrate the exploitation rates under the different hyper-parameter settings. The green line of those three figures is the result of arm $1$ which is the best arm. The red line is the result of arm $7$ which has the largest feedback expectation. Figure b shows the results that the number of event $X > 1.0$ occurs changes with the number of trials increases. The red dash line is the expectation of the ground-truth strategy. The green line is the proposed ER-UCB strategy.}
        \label{syn_problem}
    \end{figure*}
    
\section{Experiments}
    
    In the experiment section, we empirically investigate the effectiveness of the ER-UCB bandit on some synthetic and real-world AutoML tasks. Some state-of-the-art bandit strategies are selected as the compared methods, including the classical UCB (C-UCB)~\cite{bubeck2012regret}, $\epsilon$-greedy~\cite{sutton2018reinforcement}, softmax strategy~\cite{tokic2011value} and random strategy which allocates the budget by selecting arms randomly. In addition, we apply the random search on the joint hyper-parameter spaces of all algorithms (Joint) to compare with the cascaded hyper-parameter optimization.
    
    \subsection{Synthetic problem}
    
    We construct a 7-armed bandit problem in this section. The feedbacks obey Gaussian distributions with different expectations and variances: $G_1\left(0.84, 0.07^2\right)$, $G_2\left(0.84, 0.01^2\right)$, $G_3\left(0.85, 0.04^2\right)$, $G_4\left(0.85, 0.02^2\right)$, $G_5\left(0.88, 0.01^2\right)$, $G_6\left(0.88, 0.02^2\right)$, $G_7\left(0.89, 0.01^2\right)\}$. The best arm is not only related with the expectation, but also influenced by the variance.  Obviously, it is more likely to obtain the best feedback by exploiting in $G_1$, in other words, $i^*=1$. We study on the three hyper-parameters of ER-UCB firstly, and then compare the ER-UCB with other methods.
    
    \subsubsection{Hyper-parameter study}
    
    We investigate the $\theta$, $\gamma$ and $\beta$ for the ER-UCB. With fixed two of them, we study another one: with fixed $\gamma=20$, $\beta=0.85$, we study $\theta \in \left[0.0001,0.5\right]$; with fixed $\beta=0.85$, $\theta=0.01$, we study $\gamma \in \left[0,50\right]$; with fixed $\theta=0.01$, $\gamma=20$, we study $\beta \in \left[0,1.5\right]$. For every hyper-parameter, we evenly sample 1000 settings from the setting region. The core problem we care about is how the methods allocate budget to arms. Let $R^{\text{exi}}_i = \frac{T_i\left(n\right)}{n}$ define the exploitation rate for arm $i$. Large $R^{\text{exi}}_i$ means the large number of trials that the arm $i$ is selected. The trial budget is set as 1000. The experiment for every hyper-parameter setting is repeated for 3 times independently, and the average results are presented.
    
    Figure~\ref{syn_problem}:a.1, 2 and 3 show the study results of $\theta$, $\gamma$ and $\beta$. The arm $1$ is the best selection. Thus, the larger $R^{\text{exi}}_1$ the better. For $\theta$ (Figure~\ref{syn_problem}:a.1), the green line ($R^{\text{exi}}_1$) is approaching 1 when $\theta$ nears by 0. In practice, $\theta$ should be set as a small value. For $\gamma$ (Figure~\ref{syn_problem}:a.2), when $\gamma$ is small, the exploitation rates of arms are similar. And the green line is increasing during $\gamma$ is increasing. It means that the small $\gamma$ encourages exploration and the large $\gamma$ encourages exploitation according to the observations. For $\beta$ (Figure~\ref{syn_problem}:a.3), the exploitation rates are sensitive to $\beta$ when $\beta$ is around the expectations of reward distributions. Thus, $\beta$ should be carefully tuned according to different tasks. 
    
    \textbf{\begin{table}[!t]
        \centering
        \caption{The performance summary of compared bandit strategies on the synthetic problem. $\bar{X}^{*}$ is the average best feedback for three independent runnings. $i_{X^*}$ is the arm index that the best feedback is from in each of runnings. $\max_i R^{\text{exi}}_i$ is the arm index that the strategy allocates the most budget to in each of runnings. The number in bold means the best performance. }
        \label{syn_performance}
        \begin{tabular}{@{}lcccc@{}}
            \toprule
            Methods & $\bar{X}^*$             & $i_{X^{*}}$    & $\max_i R^{\text{exi}}_i$ & $R^{\text{exi}}_1$       \\ \midrule
            ER-UCB      & $\bm{1.06}$$\pm$0.02 & 1,1,1 & 1,1,1   & $\bm{0.90}$$\pm$0.01 \\
            C-UCB & 0.94$\pm$0.01 & 1,7,6 & 7,7,7   & 0.01$\pm$0.01 \\
            $\epsilon$-Greedy  & 0.98$\pm$0.04 & 6,1,1 & 6,1,6   & 0.31$\pm$0.42 \\
            Softmax     & 1.01$\pm$0.01 & 1,1,1 & 7,7,1   & 0.18$\pm$0.01 \\
            Random      & 1.00$\pm$0.05 & 1,1,1 & 4,1,6   & 0.15$\pm$0.01 \\ \bottomrule
        \end{tabular}
    \end{table}}
    
    \subsubsection{Investigation with compared methods}
    
    According to the hyper-parameter study results of the ER-UCB, we set $\theta=0.01$, $\gamma=20$, $\beta=0.85$, and compare it with the C-UCB, $\epsilon$-greedy ($\epsilon=0.1$), Softmax strategy ($\tau=0.1$) and random selection strategy. The trial budget is 1000. Every experiment is repeated for 3 times independently. The average performances are presented in Table~\ref{syn_performance}.
    
    Table~\ref{syn_performance} shows that the ER-UCB outperforms the compared methods. Furthermore, the ER-UCB can find the best arm (arm $1$) and allocate most of budget to it ($\max R^{\text{exi}}_i=\{1,1,1\}$ and average $R^{\text{exi}}_1$ is 0.9). Because the C-UCB depends only on mean observations to make decisions. It wrongly allocates budget to arm $7$ ($\max R^{\text{exi}}_i=\{7,7,7\}$). The $R^{\text{exi}}_1$ of $\epsilon$-greedy is very unstable. It means $\epsilon$-greedy can't find the best arm effectively. In general, the ER-UCB can effectively discover the best-arm and reasonably allocate budget to exploration and exploitation in this synthetic problem. 
        
    \subsection{Real-word AutoML tasks}
    
    We apply the ER-UCB to solve the real-world classification tasks. We select 10 frequently-used algorithms as the candidates from SKLEARN~\cite{sklearn-jmlr}, including DecisionTree (DT), AdaBoost (Ada), QuadraticDiscriminantAnalysis (QDA), GaussianNB (GNB), BernoulliNB (BNB), K-Neighbors (KN), ExtraTree (ET), PassiveAggressive (PA), RandomForest (RF) and SGD. And 12 classification datasets from UCI are selected as AutoML tasks. The evaluation criterion of each configuration is the accuracy score. The compared methods are C-UCB, $\epsilon$-greedy ($\epsilon=0.1$), Softmax strategy ($\tau=0.1$), random strategy and Joint. The trial budget is 1000. We set $\theta=0.01$, $\gamma=20$ for the ER-UCB on all datasets. The $\beta$ is set according to the tasks, and showed in Table~\ref{real-world-result}.  For each method and each dataset, we run every experiment 3 times independently, and the average performances of our experiment are presented. In addition, we apply the random search with 1000 trials to explore on every algorithm candidate. According to (\ref{automl-target}), we can find out the best ground-truth algorithm for the datasets. 

    \begin{table*}[!t]
        \centering
        \caption{The average performances on AutoML tasks including the best validation accuracy (V-Eval), the exploitation rate on the ground-truth best algorithm ($R^{\text{exi}}_{i^*}$), the best-selected algorithm (B. Alg.) and the test accuracy (T-Eval). The items under the dataset name are the ground-truth algorithm and the $\beta$ setting for the ER-UCB. The number in bold means the best performance in compared methods.}
        \label{real-world-result}
        \begin{tabular}{@{}clcccc|clcccc@{}}
            \toprule
            \multicolumn{1}{l}{Dataset}                   & \multicolumn{1}{c}{Methods} & V-Eval & $R^{\text{exi}}_{i^*}$     & B. Alg. & T-Eval & Dataset                    & \multicolumn{1}{c}{Methods} & V-Eval & $R^{\text{exi}}_{i^*}$     & B. Alg. & T-Eval \\ \midrule
            \multirow{6}{*}{\begin{tabular}[c]{@{}c@{}}Balance\\ (SGD) \\$\beta=0.5$\end{tabular}} & ER-UCB                      & $\bm{.9025}$   & $\bm{.5677}$ & SGD       & $\bm{.9074}$  & \multirow{6}{*}{\begin{tabular}[c]{@{}c@{}}Car\\ (ET)\\$\beta=0.6$\end{tabular}} & ER-UCB                      & $\bm{.8729}$   & $\bm{.9800}$ & ET       & $\bm{.6937}$  \\
            & C-UCB                       & .8924   & .1067 & PA       & .8339  &                            & C-UCB                       & .8690   & .1173 & RF       & .6416  \\
            & $\epsilon$-Greedy                    & .8931   & .0693 & SGD       & .8227  &                            & $\epsilon$-Greedy                    & .8630   & .0277 & RF       & .6657  \\
            & Softmax                     & .9004   & .1287 & SGD       & .8809  &                            & Softmax                     & .8620   & .1163 & DT       & .6551  \\
            & Random                      & .8978   & .1097 & SGD       & .8597  &                            & Random                      & .8628   & .1053 & RF       & .6839  \\
            \cmidrule(lr){2-6} \cmidrule(l){8-12}
            & Joint                      & .8978  & - & SGD       & .8994 &                           & Joint                      & .8619   & - & RF       & .8604  \\
            \midrule
            \multirow{6}{*}{\begin{tabular}[c]{@{}c@{}}Chess\\ (Ada)\\$\beta=0.5$\end{tabular}} & ER-UCB                      & $\bm{.9557}$   & $\bm{.6137}$ & Ada       & .8515  & \multirow{6}{*}{\begin{tabular}[c]{@{}c@{}}Cylinder\\ (ET)\\$\beta=0.5$\end{tabular}} & ER-UCB                      & $\bm{.7356}$   & $\bm{.6043}$ & ET       & $\bm{.6117}$  \\
            & C-UCB                       & .9414   & .1693 & Ada       & $\bm{.8860}$  &                            & C-UCB                       & .7172   & .1117 & ET       & .5534  \\
            & $\epsilon$-Greedy                    & .9492   & .3593 & Ada       & .7510  &                            & $\epsilon$-Greedy                    & .6528   & .0443 & ET       & .5006  \\
            & Softmax                     & .9414   & .1290 & PA       & .8229  &                            & Softmax                     & .6866   & .0977 & ET       & .5749  \\
            & Random                      & .9464   & .1070 & PA       & .6999  &                            & Random                      & .6977   & .0990 & ET       & .5534  \\
            \cmidrule(lr){2-6} \cmidrule(l){8-12}
            & Joint                      & .9457   & - & SGD       & .8479  &                            & Joint                      & .6531   & - & QDA       & .5687  \\
            \midrule
            \multirow{6}{*}{\begin{tabular}[c]{@{}c@{}}Ecoli\\ (RF)\\$\beta=0.5$\end{tabular}} & ER-UCB                      & $\bm{.8763}$   & $\bm{.7013}$ & RF       & $\bm{.8904}$  & \multirow{6}{*}{\begin{tabular}[c]{@{}c@{}}Glass\\ (DT)\\$\beta=0.4$\end{tabular}} & ER-UCB                      & $\bm{.7540}$   & $\bm{.9716}$ & RF       & $\bm{.6740}$  \\
            & C-UCB                       & .8745   & .2440 & RF       & .8333  &                            & C-UCB                       & .7265   & .1520 & RF       & .6370  \\
            & $\epsilon$-Greedy                    & .8129   & .3023 & RF       & .8809  &                            & $\epsilon$-Greedy                    & .7163   & .0017 & RF       & .6148  \\
            & Softmax                     & .8728   & .1630 & RF       & .8809  &                            & Softmax                     & .7197   & .1180 & RF       & .6148  \\
            & Random                      & .8695   & .1037 & RF       & .8762  &                            & Random                      & .7247   & .1027 & RF       & .6666  \\
            \cmidrule(lr){2-6} \cmidrule(l){8-12}
            & Joint                      & .8549   & - & RF       & .8714  &                            & Joint                      & .7087   & - & RF       & .6444  \\
            \midrule
            \multirow{6}{*}{\begin{tabular}[c]{@{}c@{}}Messider\\ (SGD) \\$\beta=0.5$\end{tabular}} & ER-UCB                      & $\bm{.7431}$   & $\bm{.5203}$ & SGD       & .7330  & \multirow{6}{*}{\begin{tabular}[c]{@{}c@{}}Nursery\\ (Ada) \\ $\beta=0.5$\end{tabular}} & ER-UCB                      & $\bm{.8200}$   & $\bm{.7640}$ & Ada       & $\bm{.6909}$  \\
            & C-UCB                       & .7297   & .1133 & SGD       & .7272  &                            & C-UCB                       & .7871   & .1370 & RF       & .6688  \\
            & $\epsilon$-Greedy                    & .7362   & .0157 & SGD       & .7272  &                            & $\epsilon$-Greedy                    & .7201   & .0607 & Ada       & .6457  \\
            & Softmax                     & .7402   & .1163 & SGD       & $\bm{.7604}$  &                            & Softmax                     & .8010   & .1150 & Ada       & .6238  \\
            & Random                      & .7406   & .1027 & SGD       & .7330  &                            & Random                      & .8039   & .1100 & Ada       & .6631  \\
            \cmidrule(lr){2-6} \cmidrule(l){8-12}
            & Joint                      & .7399   & - & SGD       & .7316  &                            & Joint                      & .7884   & - & Ada       & .6304  \\
            \midrule
            \multirow{6}{*}{\begin{tabular}[c]{@{}c@{}}Spambase\\ (Ada)\\$\beta=0.7$\end{tabular}} & ER-UCB                      & $\bm{.9328}$   & $\bm{.9853}$ & Ada       & .9449  & \multirow{6}{*}{\begin{tabular}[c]{@{}c@{}}Statlog\\ (Ada)\\$\beta=0.6$\end{tabular}} & ER-UCB                      & $\bm{.9804}$   & $\bm{.9763}$ & RF       & $\bm{.9711}$  \\
            & C-UCB                       & .9298   & .1757 & Ada       & .9457  &                            & C-UCB                       & .9779   & .1950 & RF       & .9696  \\
            & $\epsilon$-Greedy                    & .9311   & .7333 & Ada       & $\bm{.9741}$  &                            & $\epsilon$-Greedy                    & .9790   & .8047 & RF       & .9703  \\
            & Softmax                     & .9298   & .1253 & Ada       & .9471  &                            & Softmax                     & .9768   & .1397 & RF       & .9660  \\
            & Random                      & .9306   & .1057 & Ada       & .9500  &                            & Random                      & .9776   & .1203 & RF       & .9696  \\
            \cmidrule(lr){2-6} \cmidrule(l){8-12}
            & Joint                      & .9290   & - & RF       & .9428  &                            & Joint                      & .9793   & - & RF       & .9464  \\
            \midrule
            \multirow{6}{*}{\begin{tabular}[c]{@{}c@{}}WDBC\\ (Ada) \\ $\beta=0.6$\end{tabular}} & ER-UCB                      & $\bm{.9823}$   & $\bm{.9506}$ & Ada       & .9681  & \multirow{6}{*}{\begin{tabular}[c]{@{}c@{}}Wilt\\ (Ada) \\$\beta=0.8$\end{tabular}} & ER-UCB                      & $\bm{.9827}$   & $\bm{.6206}$ & RF       & $\bm{.9433}$  \\
            & C-UCB                       & .9808   & .1397 & Ada       & .9710  &                            & C-UCB                       & .9820   & .1200 & RF       & .9320  \\
            & $\epsilon$-Greedy                    & .9816   & .8757 & Ada       & .9681  &                            & $\epsilon$-Greedy                    & .9813   & .3567 & Ada       & .9427  \\
            & Softmax                     & .9794   & .1207 & Ada       & $\bm{.9739}$  &                            & Softmax                     & .9819   & .1097 & RF       & .9267  \\
            & Random                      & .9794   & .1060 & Ada       & $\bm{.9739}$  &                            & Random                      & .9821   & .1103 & RF       & .9347  \\
            \cmidrule(lr){2-6} \cmidrule(l){8-12}
            & Joint                      & .9794   & - & Ada       & .9594  &                            & Joint                      & .9821   & - & DT       & .9320  \\
            \bottomrule 
        \end{tabular}
    \end{table*}

    The average performances of the compared methods on all 12 datasets are presented in Table~\ref{real-world-result}. From Table~\ref{real-world-result}, we can get the following empirical conclusions:
    \begin{itemize}
        \item ``No free lunch'' has been proved again in those experiments. The best performance algorithms are different in different datasets. Particularly, tree-based ensemble algorithms, e.g., AdaBoost, RandomForest, etc, show the outstanding performance in most of the datasets. It indicates that the algorithm selection is necessary for making search hyper-parameters easier. 
        \item The cascaded algorithm selection and hyper-parameter optimization are necessary for making the search problem easier to solve. Comparing the random strategy with the Joint, the random strategy beats the Joint on most of the datasets (8/12). It indicates that the large search space provides more difficult for optimization.
        \item It will mislead the strategy to select wrong algorithms only according to the average performance. In Table~\ref{real-world-result}, the random strategy is not always bad in datasets. The strategies, such as C-UCB, $\epsilon$-greedy and Softmax, which focus on the average performance are easy to select wrong algorithms which average performances are good.
        \item The proposed ER-UCB bandit strategy can effectively find out the best performance algorithm (B. Alg. is the ground-truth algorithm on 9/12 datasets), and reasonably allocate the trial budget to the best algorithm (ER-UCB gets the highest $R^{\text{exi}}_{i^*}$ on 12/12 datasets). 
    \end{itemize}
    
\section{Conclusion}

    This paper proposes the extreme-region upper confidence bound (ER-UCB) bandit for the cascaded algorithm selection and hyper-parameter optimization. we employ the random search in the hyper-parameter optimization level. The level of algorithm selection is formulated as a multi-armed bandit problem. The bandit strategies are applied to allocate the limited search budget to the hyper-parameter optimization processes on algorithm candidates. However, the algorithm selection focuses on the algorithm with the maximum performance but not the average performance. To tackle this, we propose the extreme-region UCB (ER-UCB) strategy, which selects the arm with the largest extreme region of the underlying distribution. The theoretical study shows that the ER-UCB has $O\left(K \ln n\right)$ extreme-region regret upper bound, which has the same order with the classical UCB strategy. The experiments on synthetic and real-world AutoML problems empirically verify that the ER-UCB can precisely discover the algorithm with the best performance, and reasonably allocate the trial budget to the algorithm candidates.


\bibliographystyle{named}
\bibliography{erucb}

\end{document}